\documentclass[sigconf]{acmart}

\usepackage{booktabs} 

\setcopyright{rightsretained}

\usepackage{microtype}
\usepackage{graphicx}
\usepackage{subfigure}
\usepackage{booktabs} 
\usepackage{amsmath}
\usepackage{amsfonts}

\usepackage[normalem]{ulem}
\useunder{\uline}{\ul}{}

\usepackage{algorithm, algorithmic}

\usepackage{array}

\acmConference[KDD '18]{KDD Workshop on Machine Learning for Medicine and Healthcare}{August 2018}{London, UK}
\acmYear{2018}
\copyrightyear{2018}

\begin{document}
\title{Forecasting Disease Trajectories in Alzheimer's Disease Using Deep Learning}

\author{Bryan Lim}
\affiliation{%
  \institution{University of Oxford}
}
\email{bryan.lim@eng.ox.ac.uk}

\author{Mihaela van der Schaar}
\affiliation{%
  \institution{University of Oxford}
}
\affiliation{%
  \institution{Alan Turing Institute}
}
\email{mihaela.vanderschaar@eng.ox.ac.uk}

\begin{abstract}
Joint models for longitudinal and time-to-event data are commonly used in longitudinal studies to forecast disease trajectories over time. Despite the many advantages of joint modeling, the standard forms suffer from limitations that arise from a fixed model specification and computational difficulties when applied to large datasets. We adopt a deep learning approach to address these limitations, enhancing existing methods with the flexibility and scalability of deep neural networks while retaining the benefits of joint modeling. Using data from the Alzheimer's Disease Neuroimaging Institute \footnote{\label{ADNI}For the Alzheimer's Disease Neuroimaging Initiative: Data used in preparation of this article were obtained from the Alzheimer's Disease Neuroimaging Initiative (ADNI) database (adni.loni.usc.edu). As such, the investigators
within the ADNI contributed to the design and implementation of ADNI and/or provided data
but did not participate in analysis or writing of this report. A complete listing of ADNI
investigators can be found at: \url{http://adni.loni.usc.edu/wp-content/uploads/how_to_apply/ADNI_Acknowledgement_List.pdf}}, we show improvements in performance and scalability compared to traditional methods.
\end{abstract}

\maketitle

\section{Overview}
Effective clinical decision support often involves the dynamic forecasting of medical conditions based on clinically relevant variables collected over time. This involves jointly predicting the expected time to events of interest (e.g. death), biomarker trajectories, and other associated risks at different stages of disease progression. 

\begin{figure}[htb]
\includegraphics[width=1.0\linewidth]{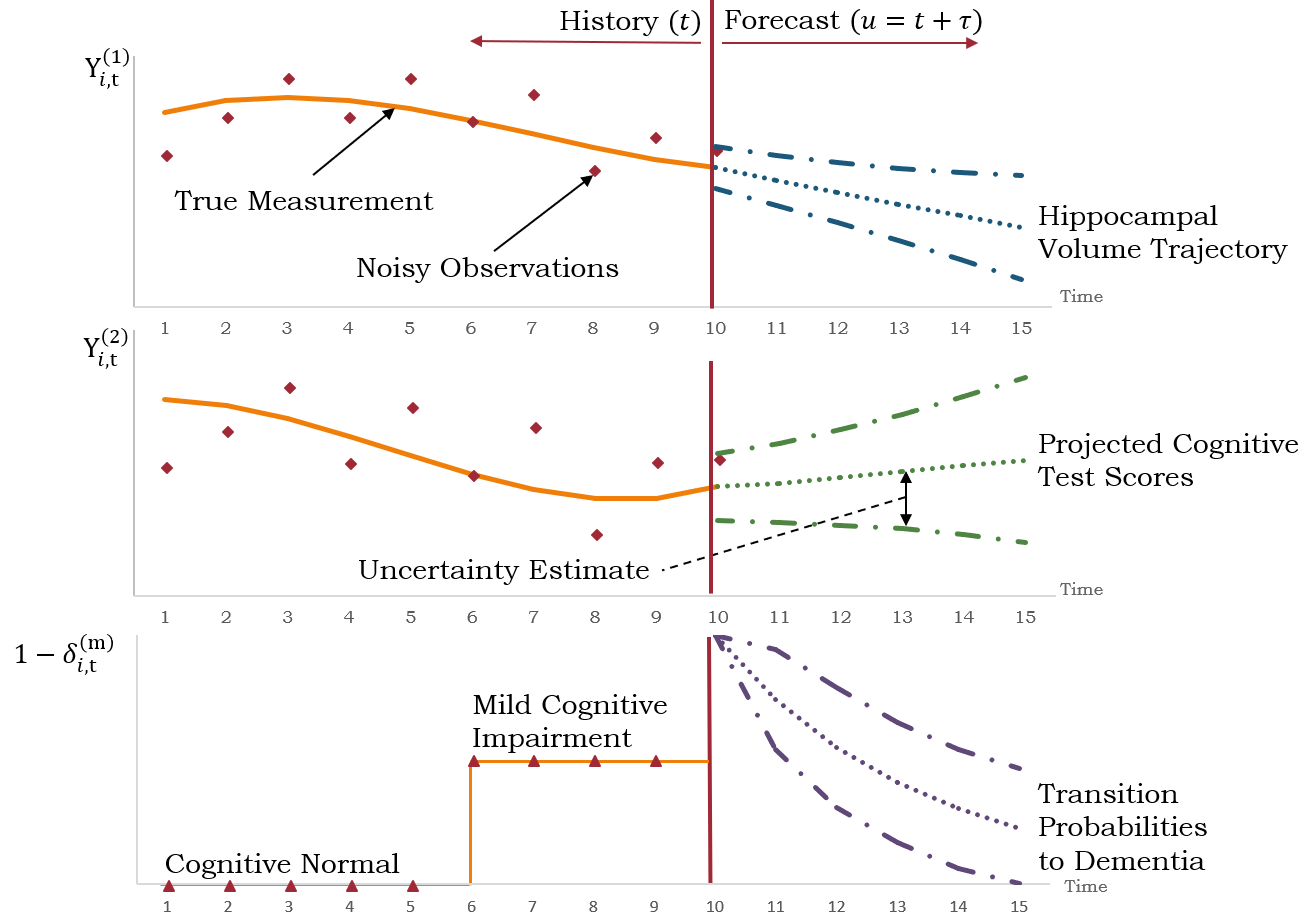}
\caption{Disease-Atlas Predictions over Time}
\label{fig:DynamicPrediction}
\end{figure}

With the prevalence of aging populations around the globe, Alzheimer's disease (AD) is a significant threat to public health - growing from being relatively rare at the start of the 20th century to having a case being reported every 7 seconds around the world \cite{ADPrevalence}. Patients at risk of developing Alzheimer's disease are usually monitored over time based on longitudinal cognitive scores and MRI measurements \cite{ADTracking}, which help doctors evaluate the severity of a patient's condition and formulate a diagnosis. As such, the ability to produce joint forecasts -- such as those depicted in Figure \ref{fig:DynamicPrediction} -- would help doctors determine both the likelihood of developing dementia and the expected rate of deterioration of a given patient, potentially allowing for intervention at an early stage.

Given the promising initial results found in our companion paper for Cystic Fibrosis patients \cite{DiseaseAtlas}, we investigate the application of the Disease-Atlas -- a novel conception of the joint modeling framework using deep learning -- to jointly predicting the expected time-to-transition to Alzheimer's disease and the values of longitudinal measurements, providing additional clinical decision support to doctors evaluating potential Alzheimer's disease patients. We start with an overview of the Disease-Atlas in Section  \ref{sec:probdefinition} and \ref{sec:architecture}, demonstrating performance gains for tests on data from the ADNI in Section \ref{sec:ADTests}.

\section{Problem Definition}
\label{sec:probdefinition}
For a given longitudinal study, let there be $N$ patients with observations made at time $t$, for $ 0 \le t \le T_{cens}$ where $T_{cens}$ denotes an administrative censoring time \footnote{\label{administrativecensoring} Administrative censoring refers to the right-censoring that occurs when a study observation period ends. }. For the $i^{th}$ patient at time $t$, observations are made for a $K$-dimensional vector of longitudinal variables $\mathbf{V_{i,t}} = [Y_{i,t}^{(1)}, \dots, Y_{i,t}^{(C)}, B_{i,t}^{(1)}, \dots, B_{i,t}^{(D)} ]$, where $Y_{i,t}^{(c)}$ and $B_{i,t}^{(d)}$ are continuous and discrete longitudinal measurements respectively, a $L$-dimensional vector of external covariates $\mathbf{X_{i,t}} = [X_{i,t}^{(1)}, \dots X_{i,t}^{(L)} ]$, and a $M$-dimensional vector of event occurrences $\mathbf{\delta_{i,t}} = [\delta_{i,t}^{(1)}, \dots, \delta_{i,t}^{(M)} ]$ , where $\delta_{i,t}^{(m)} \in \{0, 1\}$ is an indicator variable denoting the presence or absence of the $m^{th}$ event. $T_{i,t}^{(m)}$ is defined to be the first time the event is observed after $t$, which allows us to model both repeated events and events that lead to censoring (e.g. death). The final observation for patient i occurs at $T_{i,\max} =  \min( T_{cens}, T_{i,0}^{(a_1)}, \dots , T_{i,0}^{(a_{\max})})$, where $ \{ a_i, \dots, a_{\max} \} $ is the set of indices for events that censor observations.

\section{Disease-Atlas Architecture}
\label{sec:architecture}
\begin{figure*}[htp]
\centering
\noindent\makebox[\textwidth][c]{
\includegraphics[width=1.05\linewidth]{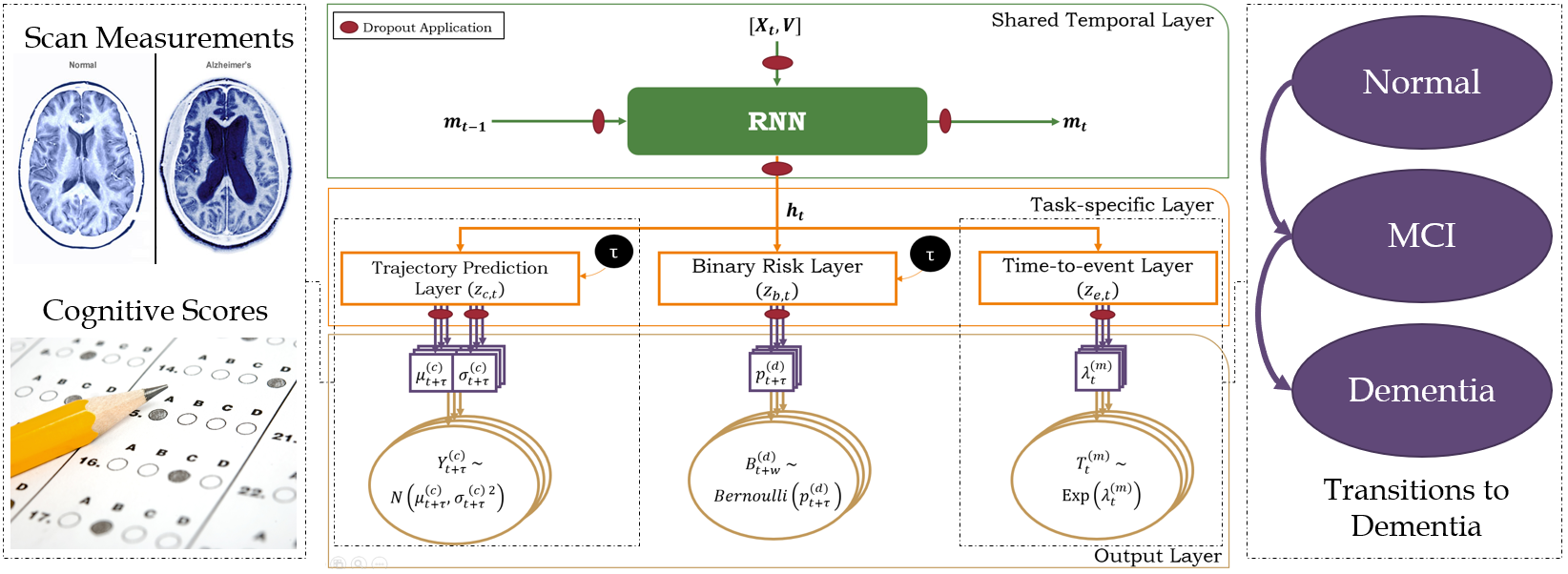}
}
\caption{Disease-Atlas Network Architecture}
\label{fig:NetworkArchitecture}
\end{figure*}

The Disease-Atlas captures the associations within the joint modeling framework, by learning \textit{shared representations} between trajectories at different stages of the network, while retaining the same sub-model distributions captured by joint models. The network, as shown in Figure \ref{fig:NetworkArchitecture}, is conceptually divided into 3 sections: 1) A shared temporal layer to learn the temporal and cross-correlations between variables, 2) task-specific layers to learn shared representations between related trajectories, and 3) an output layer which computes parameters for predictive sub-model distributions for use in likelihood loss computations during training and generating predictive distributions at run-time.

The equations for each layer are listed in detail below. For notational convenience, we drop the subscript $i$ for variables in this section, noting that the network is only applied to trajectories from one patient at time. While we focus on both the continuous valued and time-to-event predictions for tests in Section \ref{sec:ADTests}, we include descriptions of binary predictions here for completeness.
\paragraph{Shared Temporal Layer}
We start with an RNN at the base of the network, which incorporates historical information into forecasts by updating its memory state over time. For the tests in Section \ref{sec:ADTests},  we adopt the use of a long-short term memory network (LSTM) in the base layer. 
\begin{align}
[ \mathbf{h_t}, \mathbf{m_t} ] = \text{RNN}(  [\mathbf{X_{t}}, \mathbf{V_{t}}] , \mathbf{m_{t-1}})&
\end{align}
Where $\mathbf{h_t}$ is the output of the RNN and $\mathbf{m_t}$ its memory state. To generate uncertainty estimates for forecasts and retain consistency with joint models, we adopt the MC dropout approach described in \citep{GalDropoutRNN}. Dropout masks are applied to the inputs, memory states and outputs of the RNN, and are also fixed across time steps. For memory updates, the RNN uses the Exponential Linear Unit (ELU) activation function.
\paragraph{Task-specific Layers}
For the task-specific layers, variables can be grouped according to the types of outputs, with layer $\mathbf{z_{c,t}}$ for continuous-valued longitudinal variables, $\mathbf{z_{b,t}}$ for binary longitudinal variables and $\mathbf{z_{e,t}}$ for events. Dropout masks are also applied to the outputs of each layer here. At the inputs to the continuous and binary task layers, a prediction horizon $\tau$ is also concatenated with the outputs from the RNN. This allows the parameters of the predictive distributions at $t+\tau$ to be computed in the final layer, i.e. $\mathbf{ \tilde{h}_{t}} = \left[ \mathbf{h}_t , \tau  \right]$.
\begin{subequations}
\begin{align}
\mathbf{z_{c,t}} = \text{ELU}( \mathbf{W_c} \mathbf{\tilde{h}_t} +  \mathbf{a_c})\\
\mathbf{z_{b,t}} = \text{ELU}( \mathbf{W_b} \mathbf{\tilde{h}_t} +  \mathbf{a_b})\\
\mathbf{z_{e,t}} = \text{ELU}( \mathbf{W_e} \mathbf{h_t} +  \mathbf{a_e})
\end{align}
\end{subequations}
\paragraph{Output Layer}
The final layer computes the parameter vectors of the predictive distribution, which are used to compute log likelihoods during training and dynamic predictions at run-time.
\begin{subequations}
\begin{align}
\mathbf{\mu_{t+\tau}} &=  \mathbf{W_{\mu}} \mathbf{z_{c,t}} +  \mathbf{a_{\mu}}  \\
\mathbf{\sigma_{t+\tau}}& = \text{Softplus}( \mathbf{W_{\sigma}} \mathbf{ z_{c,t}} +  \mathbf{a_{\sigma}})) \\
\mathbf{p_{t+\tau}}& = \text{Sigmoid}( \mathbf{W_{p}} \mathbf{ z_{b,t}} +  \mathbf{a_{p}})) \\
\lambda_{t} & = \text{Softplus}( \mathbf{W_{\lambda}} \mathbf{ z_{e,t}} +  \mathbf{a_{\lambda}}))  
\end{align}
\end{subequations}
Softplus activation functions are applied to $\mathbf{\sigma_{t+\tau}}$ and $\mathbf{p_{t+\tau}}$ to ensure that we obtain valid (i.e. $\ge 0$) standard deviations and binary probabilities. For simplicity, the exponential distribution is selected to model survival times, and predictive distributions are as below:
\begin{subequations}
\label{eqn:submodeldist}
\begin{align}
Y_{t+\tau}^{(c)} &\sim N\left(\mu_{t+\tau}^{(c)}, \sigma_{t+\tau}^{ (c)  2}\right)\\
B_{t+\tau}^{(d)} &\sim \text{Bernoulli} \left(p_{t+\tau} ^{(d)} \right)\\
T_t^{(m)} &\sim \text{Exponential} \left( \lambda_{t}^{(m)} \right)
\end{align}
\end{subequations}
\subsection{Multitask Learning}
\label{sec:MultitaskLearning}
From the above, the negative log-likelihood of the data given the network is:
\begin{align}
\label{eqn:multivariateloss}
\mathcal{L}(\mathbf{W}) = 
\sum_{i,t,w,k_c,k_b,m} - \biggl[& \log f_c\left(Y_{i,t+\tau}^{(c)} |\mu_{t+\tau}^{(c)}, \sigma_{t+\tau}^{(c) 2},  \mathbf{W} \right) \\
&+ \log f_b\left(B_{i,t+\tau}^{(d)} |p_{t+\tau} ^{(d)}, \mathbf{W} \right) & \nonumber\\
& + \log f_T \left(T_{i,t}^{(m)} | \lambda_{t}^{(m)}, \mathbf{W} \right)\biggr]&
\end{align} 
Where $f_c(.), f_b(.)$ are likelihood functions based on Equations \ref{eqn:submodeldist} and $\mathbf{W}$ collectively represents the weights and biases of the entire network. For survival times, $f_T(.)$ is given as:
\begin{equation}
f_T \left(T_t^{(m)} | \lambda_{t}^{(m)}, \mathbf{W} \right)  = \left(\lambda_{t}^{(m)} \right)^{\delta_{i,T}} \exp\left(-\lambda_{t}^{(m)} T_t^{(m)}\right)
\end{equation} 
Which corresponds to event-free survival until time T before encountering the event \citep{generalizedLinearModels}. While the negative log-likelihood can be directly optimized across tasks, the use of multitask learning can yield the following benefits:
\paragraph{Better Survival Representations} As shown in \citep{ADMultitaskClassification}, multitask learning problems which have one main task of interest can weight the individual loss contributions of each subtask to favor representations for the main problem. For our current architecture, where we group similar tasks into task-specific layers, our loss function corresponds to:
\begin{align}
\label{eqn:multioutputloss}
L(\mathbf{W}) = &- \underbrace{\alpha_c \sum^{i,t,w,c} \log f_c\left(Y_{t+\tau}^{(c)} | \mathbf{W} \right) }_{\text{Continuous Longitudinal Loss } l_c } 
- \underbrace{\alpha_b \sum^{i,t,w,d}  \log f_b\left(B_{t+\tau}^{(d)} | \mathbf{W} \right)}_{\text{Binary Longitudinal Loss } l_b} & \nonumber\\
 &- \underbrace{\alpha_T \sum^{i,t,m} \log f_T \left(T_{t}^{(m)} | \mathbf{W} \right)}_{\text{Time-to-event Loss } l_T } &
\end{align} 
Given that survival predictions are the primary focus of many longitudinal studies, we set $\alpha_c = \alpha_b = 1$ and include $\alpha_T$ as an additional hyperparameter to be optimized. To train the network, patient trajectories are subdivided into Q sets of $\Omega_q(i, \rho, \tau) = \left\lbrace \mathbf{X_{i,{0:\rho}}}, \mathbf{Y_{i, \rho+\tau}}, \mathbf{T_{max,i}}, \mathbf{\delta_i} \right\rbrace$, where $\rho$ is the length of the covariate history to use in training trajectories up to a maximum of $\rho_{\max}$. Full details on the procedure can be found in Algorithm \ref{alg:multitasktraining}. 
\paragraph{Handling Irregularly Sampled Data} We address issues with irregular sampling by grouping variables that are measured together into the same task, and training the network with multitask learning. For instance, volumes of different parts of the brain (e.g. hippocampal, ventricular and intra-cranial volume) that are be measured together during the same MRI scan session can be grouped together in the same task. Given the completeness of the datasets we consider, we assume that task groupings match those defined by the task-specific layer of the network, and multitask learning is performed using Equation \ref{eqn:multioutputloss} and Algorithm \ref{alg:multitasktraining}. 

We note, however, that in the extreme case where none of the trajectories are aligned, we can define each variable as a separate task with its own loss function $l_*$. Algorithm \ref{alg:multitasktraining} then samples loss functions for one variable at a time, and the network is trained using only actual observations as target labels. This could reduce errors in cases where multiple sample rates exist and simple imputation is used, which might result in the multioutput networks replicating the imputation process instead of making true predictions. 

\begin{algorithm}[tbp]
   \caption{Training Disease-Atlas}
   \label{alg:multitasktraining}
\begin{algorithmic}
   \STATE {\bfseries Input:} Data $\Omega=\{\Omega_1, \dots, \Omega_Q \}$, max iterations $\mathcal{J}$
   \STATE {\bfseries Output:} Calibrated network weights $\mathbf{W}$
   \FOR{ \texttt{count}$=1$ {\bfseries to} $\mathcal{J}$}
   \STATE Get minibatch $\mathcal{M} \sim$  $\gamma$ random samples from $\Omega$
   \STATE Sample task loss function $l \sim \{l_c, l_b, l_T\}$ 
   \STATE Update $\mathbf{W} \leftarrow \texttt{Adam}(l, \mathcal{M})$, using feed-forward passes with  dropout applied
   \ENDFOR
\end{algorithmic}
\end{algorithm}
\subsection{Forecasting Disease Trajectories}
\label{sec:dropout}
Dynamic prediction involves 2 key elements - 1) calculating the expected longitudinal values and survival curves as described above, and 2) computing uncertainty estimates. To obtain these measures, we apply the Monte-Carlo dropout approach of \citep{GalDropoutRNN} by approximating the posterior over network weights as:
\begin{equation}
p(V^{(k)}_{t+\tau} | \mathcal{F}_t) \approx \frac{1}{J} \sum_{j=1}^J p(V^{(k)}_{t+\tau} | \mathcal{F}_t, \hat{\mathbf{W}}_j )
\end{equation}
Where we draw  $J$ samples $\hat{\mathbf{W}}_j$ using feed-forward passes through the network with the same dropout mask applied across time-steps. The samples obtained can then be used to compute expectations and uncertainty intervals for forecasts. 

\section{Performance Evaluation For Alzheimer's Disease}
\label{sec:ADTests}
\subsection{Data Description}
The Alzheimer's Disease Neuroimaging Initiative (ADNI) study data is a comprehensive dataset that tracks the progression of the Alzheimer's disease (AD) through 3 main states: normal brain function, mild cognitive impairment and the onset of either the disease or dementia. This data surveys 1737 patients for periods up to 10 years, capturing informative features extracted with Positron Emission Tomography (PET) regions of interest (ROI) scans -- e.g. measures of cell metabolism, which are known to be reduced for AD patients -- Magnetic Resonance (MRI) and Diffusion Tensor imaging (DTI) (for instance, ventricles volume), CSF and blood biomarkers, genetics, cognitive tests (ADAS-Cog), demographic and others. Observations were discretized to 6-month (or 0.5 year) intervals, and missing measurements were imputed using the previous value if present, and the population mean otherwise. In this investigation, we use a random selection of $60\%$ of patients for our training data, $20\%$ for validation and the final $20\%$ for evaluation as per the CF tests. This was repeated 3 times to form 3 different partitions of the dataset, which were then used for cross-validation. The Disease-Atlas was used to jointly forecast longitudinal observations of clinical scores and scan measurements, treating the transition to Alzheimer's Disease from either mild cognitive impairment (MCI) or Cognitively Normal (CN) states as our event of interest. Hyperparameter optimization was performed with 20 iterations of random search.

\subsection{Results \& Discussion}
To evaluate predictions of the event-of-interest -- i.e. transitions to dementia -- we compared the performance of the Disease-Atlas against simpler recurrent neural networks (i.e. LSTMs) and standard methods from biostatistics (i.e. landmarking \cite{vanHouwelingenDynamicPrediction} and joint models (JM) fitted with a two-step approximation \cite{MixedEffectsModelsForComplexData}). 

Prediction results for transitions to dementia -- in terms of the area under the receiver operating characteristic (AUROC) and the precision-recall curve (AUPRC) -- and MSE improvements for longitudinal forecasts can be found in Tables \ref{tab:ADsurvival} and \ref{tab:ADMSE} respectively. From the cross-validation performance, we see that the Disease-Atlas consistently outperforms both the standard neural network and traditional benchmarks for survival analysis particularly on short-term horizons -- improving the LSTM by $10\%$ and JM by $7\%$ on average across all time steps. 

For longitudinal predictions, we focus on both the Disease-Atlas and JM which are able to generate predictions at arbitrary time steps in the future. Once again, the Disease-Atlas outperforms joint models across the majority of longitudinal variables and time steps, with gains of $40\%$ on average -- highlighting the benefits of a deep learning approach to joint modeling.

\section{Conclusions}
In this paper, we investigate an application of the Disease-Atlas to forecasting longitudinal measurements and expected time-to-transition to Dementia for patients at risk of Alzheimer's Disease. Using data from the ADNI, the Disease-Atlas \cite{DiseaseAtlas} demonstrated performance gains over both simpler neural networks such as LSTMs and traditional methods from biostatistics -- demonstrating the advantages of the Disease-Atlas as a method for joint modeling and highlighting its potential as a tool for clinical decision support.

\begin{table}[htb]
\centering
\caption{Cross-Validation Performance for Transitions to Alzheimer's Disease (Mean $\pm$ S.D.)}
\label{tab:ADsurvival}
\begin{tabular}{@{}l|l|ll@{}}
\toprule
\textbf{}      & \textbf{$\tau$} & \textbf{Disease-Atlas}       & \textbf{LSTM}                \\ \midrule
\textbf{AUROC} & 0.5             & \textbf{0.954 ($\pm$ 0.008)} & 0.938 ($\pm$ 0.005)          \\
               & 1               & \textbf{0.935 ($\pm$ 0.006)} & 0.929 ($\pm$ 0.005)          \\
               & 1.5             & \textbf{0.906 ($\pm$ 0.004)} & 0.905 ($\pm$ 0.001)          \\
               & 2               & \textbf{0.899 ($\pm$ 0.014)} & \textbf{0.899 ($\pm$ 0.006)} \\ \cmidrule(l){2-4} 
               & \textbf{$\tau$} & \textbf{Landmarking}         & \textbf{JM}                  \\ \cmidrule(l){2-4} 
               & 0.5             & 0.913 ($\pm$ 0.033)          & 0.916 ($\pm$ 0.035)          \\
               & 1               & 0.914 ($\pm$ 0.010)          & 0.919 ($\pm$ 0.016)          \\
               & 1.5             & 0.892 ($\pm$ 0.012)          & 0.897 ($\pm$ 0.007)          \\
               & 2               & 0.884 ($\pm$ 0.023)          & 0.890 ($\pm$ 0.015)          \\ \midrule
               & \textbf{$\tau$} & \textbf{Disease-Atlas}       & \textbf{LSTM}                \\ \midrule
\textbf{AUPRC} & 0.5             & \textbf{0.326 ($\pm$ 0.038)} & 0.256 ($\pm$ 0.036)          \\
               & 1               & \textbf{0.271 ($\pm$ 0.043)} & 0.268 ($\pm$ 0.015)          \\
               & 1.5             & \textbf{0.211 ($\pm$ 0.043)} & 0.198 ($\pm$ 0.018)          \\
               & 2               & \textbf{0.183 ($\pm$ 0.056)} & 0.178 ($\pm$ 0.027)          \\ \cmidrule(l){2-4} 
               & \textbf{$\tau$} & \textbf{Landmarking}         & \textbf{JM}                  \\ \cmidrule(l){2-4} 
               & 0.5             & 0.270 ($\pm$ 0.048)          & 0.295 ($\pm$ 0.078)          \\
               & 1               & 0.240 ($\pm$ 0.056)          & 0.257 ($\pm$ 0.083)          \\
               & 1.5             & 0.174 ($\pm$ 0.040)          & 0.185 ($\pm$ 0.040)          \\
               & 2               & 0.167 ($\pm$ 0.050)          & \textbf{0.183 ($\pm$ 0.045)} \\ \bottomrule
\end{tabular}
\end{table}
\begin{table}[htb]
\centering
\caption{\% Decrease in MSE for Longitudinal Predictions between Disease-Atlas \& JM (Mean $\pm$ S.D.)}
\label{tab:ADMSE}
\centerline{
\begin{tabular}{@{}lllll@{}}
\toprule
\textbf{$\tau$  (Years)} & \textbf{0.5}             & \textbf{1}       & \textbf{1.5}     & \textbf{2}       \\ \midrule
{\ul \textbf{MRI}}       &                          &                  &                  &                  \\
ICV                      & 63\% ($\pm$8\%)          & 62\% ($\pm$8\%)  & 62\% ($\pm$7\%)  & 61\% ($\pm$6\%)  \\
WholeBrain               & 51\% ($\pm$9\%)          & 51\% ($\pm$10\%) & 51\% ($\pm$10\%) & 51\% ($\pm$8\%)  \\
Ventricles               & 78\% ($\pm$7\%)          & 77\% ($\pm$7\%)  & 77\% ($\pm$6\%)  & 76\% ($\pm$6\%)  \\
Hippocampus              & 63\% ($\pm$0\%)          & 63\% ($\pm$0\%)  & 62\% ($\pm$1\%)  & 62\% ($\pm$1\%)  \\
Fusiform                 & 61\% ($\pm$12\%)         & 59\% ($\pm$12\%) & 57\% ($\pm$12\%) & 56\% ($\pm$11\%) \\
MidTemp                  & 64\% ($\pm$5\%)          & 62\% ($\pm$5\%)  & 61\% ($\pm$4\%)  & 59\% ($\pm$4\%)  \\
Entorhinal               & 56\% ($\pm$9\%)          & 54\% ($\pm$7\%)  & 52\% ($\pm$6\%)  & 50\% ($\pm$6\%)  \\ \midrule
{\ul \textbf{Cognitive}} &                          &                  &                  &                  \\
CDRSB                    & 34\% ($\pm$4\%)          & 30\% ($\pm$2\%)  & 27\% ($\pm$2\%)  & 25\% ($\pm$2\%)  \\
MMSE                     & 18\% ($\pm$3\%)          & 16\% ($\pm$1\%)  & 16\% ($\pm$1\%)  & 17\% ($\pm$2\%)  \\
ADAS11                   & \textit{-8\% ($\pm$6\%)} & 9\% ($\pm$5\%)   & 15\% ($\pm$6\%)  & 19\% ($\pm$5\%)  \\
RAVLT Imm.               & 39\% ($\pm$4\%)          & 36\% ($\pm$2\%)  & 33\% ($\pm$2\%)  & 31\% ($\pm$2\%)  \\
RAVLT Learn.             & 21\% ($\pm$1\%)          & 20\% ($\pm$1\%)  & 19\% ($\pm$2\%)  & 18\% ($\pm$3\%)  \\
ADAS13                   & \textit{0\% ($\pm$11\%)} & 14\% ($\pm$10\%) & 19\% ($\pm$9\%)  & 22\% ($\pm$10\%) \\
RAVLT Forget.            & 23\% ($\pm$4\%)          & 32\% ($\pm$5\%)  & 32\% ($\pm$5\%)  & 33\% ($\pm$6\%)  \\
RAVLT Perc. F.           & 13\% ($\pm$5\%)          & 29\% ($\pm$4\%)  & 32\% ($\pm$3\%)  & 35\% ($\pm$0\%)  \\ \bottomrule
\end{tabular}
}
\end{table}
\pagebreak
\bibliographystyle{ACM-Reference-Format}
\bibliography{disease_atlas}

\end{document}